\documentclass[10pt,twocolumn,letterpaper]{article}

\usepackage[pagenumbers]{cvpr} 

\usepackage[table]{xcolor}
\usepackage{graphicx}
\usepackage{amsmath}
\usepackage{amssymb}
\usepackage{mathtools}
\usepackage{threeparttable}
\usepackage{tabularx}
\usepackage{multirow}
\usepackage{booktabs}
\usepackage{cellspace}
\usepackage{color}
\usepackage{caption}
\usepackage{hhline}
\usepackage{xfrac}
\usepackage{makecell}
\usepackage{adjustbox}
\usepackage{bm}
\usepackage{enumerate}
\usepackage{arydshln}
\usepackage{algorithmic}
\usepackage{algorithm}
\usepackage{amsthm}

\usepackage{etoolbox} 

\makeatletter 
\patchcmd{\maketitle}%
  {\def\@makefnmark{\rlap{\@textsuperscript{\normalfont\@thefnmark}}}}%
  {\def\@makefnmark{\rlap{\@textsuperscript{\normalfont\color{black}\@thefnmark}}}}%
  {}
  {}
\makeatother 

\usepackage[pagebackref,breaklinks,colorlinks]{hyperref}
\usepackage[symbol]{footmisc}

\definecolor{best}{RGB}{255, 153, 153}
\definecolor{second}{RGB}{255, 204, 153}
\newcolumntype{C}{>{\centering\arraybackslash}X}

\usepackage[capitalize]{cleveref}
\crefname{section}{Sec.}{Secs.}
\Crefname{section}{Section}{Sections}
\Crefname{table}{Table}{Tables}
\crefname{table}{Tab.}{Tabs.}

\def\cvprPaperID{*****} 
\def\confName{CVPR}
\def\confYear{2023}

\begin{document}

\title{Semantic-aware Occlusion Filtering Neural Radiance Fields in the Wild}

\author{Jaewon Lee$^{1}$ ~~~ Injae Kim$^1$ ~~~ Hwan Heo$^1$ ~~~ Hyunwoo J. Kim$^{1}$\textcolor{black}{\thanks{corresponding author.}}\\
{\small ${}^1$Korea University}\\
{\tt\small \{2j1ejyu, dna9041, gjghks950, hyunwoojkim\}@korea.ac.kr}
}

\maketitle

\def\eg{\emph{e.g}\onedot}
\def\ie{\emph{i.e}\onedot}
\newcommand{\paragrapht}[1]{\noindent\textbf{#1}}

\begin{abstract}
We present a learning framework for reconstructing neural scene representations from a small number of unconstrained tourist photos.
Since each image contains transient occluders, decomposing the static and transient components is necessary to construct radiance fields with such in-the-wild photographs where existing methods require a lot of training data.
We introduce SF-NeRF, aiming to disentangle those two components with only a few images given, which exploits semantic information without any supervision.
The proposed method contains an occlusion filtering module that predicts the transient color and its opacity for each pixel, which enables the NeRF model to solely learn the static scene representation.
This filtering module learns the transient phenomena guided by pixel-wise semantic features obtained by a trainable image encoder that can be trained across multiple scenes to learn the prior of transient objects.
Furthermore, we present two techniques to prevent ambiguous decomposition and noisy results of the filtering module.
We demonstrate that our method outperforms state-of-the-art novel view synthesis methods on Phototourism dataset in a few-shot setting.
\end{abstract}

\section{Introduction}
\label{sec:1}
In recent years, synthesizing novel views from 2D images has received increasing attention due to the rapid development of neural rendering techniques.
Particularly, neural radiance fields (NeRF)~\cite{mildenhall2020nerf} has shown remarkable performance on novel view synthesis by implicitly encoding volumetric density and color of a 3D scene through a multi-layer perceptron (MLP).
With the success of NeRF, several subsequent works have been proposed to extend the neural field to speed up training and rendering~\cite{mueller2022instant,barron2021mipnerf,yu2021plenoctrees,yu2021plenoxels,garbin2021fastnerf,reiser2021kilonerf}, handle dynamic scenes~\cite{song2022nerfplayer,pumarola2020dnerf,park2021nerfies,park2021hypernerf,gao2021dynamic,li2021neural,tretschk2021non,du2021neural,yuan2021star}, learn scene representation with few images~\cite{kim2022infonerf,yu2021pixelnerf,2022sinnerf,jain2021dietnerf,Niemeyer2021Regnerf,muller2022autorf,liu2022neuray,rematas2021sharf,wang2021ibrnet,wei2021nerfingmvs,johari2022geonerf} and so on.
However, most of these approaches have been demonstrated in controlled settings, where the radiance of the scene across all images does not change and every content in the scene is static.
In the case of real-world images (\emph{e.g.}, internet photographs of cultural landmarks), they don't follow this assumption: the illumination varies depending on the time and weather the photo was taken, and moving objects such as clouds, people or cars could appear.

A number of studies have conducted to handle the photometric variation and transient objects.
Previous works have mainly addressed the inconsistent appearance by using an appearance embedding for each image and optimizing them~\cite{martinbrualla2020nerfw,tancik2022blocknerf,turki2022mega}.
As for the transient phenomena which we will focus on, NeRF-W~\cite{martinbrualla2020nerfw} and HA-NeRF~\cite{chen2022hanerf} utilize an additional transient module that separates transient components from the scene.
On the other hand, Block-NeRF~\cite{tancik2022blocknerf} and Mega-NeRF~\cite{turki2022mega} use segmentation models to mask out objects of classes that are generally considered as movable objects.
However, the former requires a large number of images since the model has to get rid of the occluders as well as to learn the complicated geometry and appearance of the scene, and the latter is restricted to predefined classes, which could miss exceptional objects such as shadows.

To address these limitations, we propose a novel framework named SF-NeRF, which utilizes two additional modules: an image encoder that learns the prior of transient occluders and an occlusion filtering module called FilterNet.
FilterNet decomposes static and transient phenomena by predicting the transient components: color and opacity for each image, conditioned on the image feature given by the encoder.
In contrast to the previous methods above, our method is not constrained to predefined classes and can learn features that can be generally used to disentangle transient objects from the scene, which enables few-shot learning.
Moreover, we apply a reparameterization technique to the transient opacity by modeling it as a Binary Concrete random variable to separate transient objects from the scene completely.
We also add a regularization term to impose smoothness constrain on the transient opacity.

We evaluate our method on Phototourism dataset~\cite{jin2021phototourism}, which includes internet photos taken at cultural landmarks, under the few-shot setting, training with 30 images for each landmark.
Our experiments demonstrate that SF-NeRF can learn the prior of transient occluders and thus decompose the scene with only a few training images.

In summary, our contributions can be summarized as follows:
\begin{itemize}
    \item[\textbullet] We propose a novel framework (SF-NeRF) that learns to decompose static and transient components of the scene by exploiting semantic features of the images in an unsupervised manner.
    \item[\textbullet] We introduce a reparameterization technique within FilterNet during training to avoid ambiguous decomposition of the scene.
    \item[\textbullet] We introduce a regularization term to ensure the smoothness of transient opacity field.
    \item[\textbullet] The proposed method outperforms state-of-the-art novel view synthesis methods on Phototourism dataset under the few-shot setting.
\end{itemize}

\section{Related Works}

\subsection{Neural Rendering}
In recent years, neural scene representations have been extensively studied to achieve novel view synthesis and 3D reconstruction tasks~\cite{mildenhall2020nerf,bi2020deep,lombardi2019neural,thies2019deferred,zhou2018stereo,sitzmann2019deepvoxels,dai2020pointcloudneural,mescheder2019occupancynetworks,park2019deepsdf}.
In particular, neural radiance fields (NeRF)~\cite{mildenhall2020nerf} represents a 3D scene by combining differentiable volume rendering with MLPs and achieves photorealistic novel view synthesis.
After the great success of the NeRF, several subsequent works have attempted to improve its performance on view synthesis~\cite{barron2021mipnerf,mildenhall2022nerf,verbin2022ref} or extend it to solve other neural rendering tasks such as generative tasks~\cite{2021GRAF,Niemeyer2020GIRAFFE,jain2021dreamfields,poole2022dreamfusion,chan2021pigan}, fast rendering~\cite{liu2020nsvf,reiser2021kilonerf,yu2021plenoctrees,yu2021plenoxels,garbin2021fastnerf,mueller2022instant}, few-shot view synthesis~\cite{kim2022infonerf,Niemeyer2021Regnerf,jain2021dietnerf,wang2021ibrnet,muller2022autorf,liu2022neuray,chen2021mvsnerf,yu2021pixelnerf,johari2022geonerf}, pose estimation~\cite{wang2021nerfmm,2020inerf,SCNeRF2021,lin2021barf}, dynamic view synthesis~\cite{park2021nerfies,park2021hypernerf,pumarola2020dnerf,song2022nerfplayer,gao2021dynamic,li2021neural,tretschk2021non,du2021neural,yuan2021star}, relighting~\cite{boss2021nerd,Pratul2021nerv,bi2020neural,chen2020neural} and so on.
Martin-Brualla \emph{et al}.~\cite{martinbrualla2020nerfw} and Chen \emph{et al}.~\cite{chen2022hanerf} aim to solve view synthesis with internet photo collections which contain transient occluding subjects and variable illumination.
Our work focus on training with these photos in a few-shot setting.

\subsection{3D scene decomposition}
Research towards scene decomposition has been conducted for various purposes.
A number of studies separate foreground and background NeRF models in order to handle unbounded scenes for outdoor environments~\cite{2020nerfpp, turki2022mega, rematas2022urban}.
In the case of dynamic scenes, several works decompose static NeRF model for the background and dynamic NeRF for the foreground, which includes objects that move within the scene~\cite{ost2021neural, gao2021dynamic, xie2021fignerf, li2021neural, tretschk2021non}.
Some studies aim to decompose scenes consisting of multiple objects~\cite{benaim2022volumetric, kobayashi2022dff, yang2021learning, wang2022dmnerf, stelzner2021obsurf, yu2021uorf, smith2022unsupervised, tancik2022blocknerf, turki2022mega, rematas2022urban} and further manipulate those objects separately~\cite{benaim2022volumetric, kobayashi2022dff, yang2021learning, wang2022dmnerf} with the help of external knowledge such as segmentation features~\cite{kobayashi2022dff} or masks~\cite{benaim2022volumetric, yang2021learning, wang2022dmnerf, tancik2022blocknerf, turki2022mega, rematas2022urban}.

Reconstructing 3D scenes from photo collections taken in real-world environments such as the Phototourism dataset~\cite{jin2021phototourism} is challenging since the scene is occluded by various objects in most images.
Therefore, disentangling the transient elements from the static scene is essential. 
NeRF-W~\cite{martinbrualla2020nerfw} address this by adding a "transient" head to the NeRF model that emits its own color and density conditioned on image-wise embeddings.
Instead of reconstructing transient objects by using a 3D transient field, HA-NeRF~\cite{chen2022hanerf} replace it with an image-dependent 2D visibility map which masks out the transient part of the image.
However, NeRF-W~\cite{martinbrualla2020nerfw} and HA-NeRF~\cite{chen2022hanerf} cannot decompose static and transient components when a small number of training images are given.
We aim to focus on disentangling transient components from the scene with few images.

\section{Methods}
\label{sec:4}
We propose a novel learning framework, named as ``\textbf{S}emantic-aware Occlusion \textbf{F}iltering \textbf{Ne}ural \textbf{R}adiance \textbf{F}ields (SF-NeRF)", to learn NeRF representations from only a few in-the-wild photographs.
In order to decompose static and transient components consistently, we introduce a new occlusion handling module called FilterNet, which predicts the transient components with the help of semantic information of the images extracted by an unsupervised pre-trained encoder in Sec.~\ref{sec:4.1}.
We then use one reparameterization trick to avoid ambiguous decomposition in Sec.~\ref{sec:4.2}.
During training, we employ a prior to prevent FilterNet's noisy results in Sec.~\ref{sec:4.3}.
Our overall pipeline is illustrated in Figure~\ref{fig:pipeline}. 

\subsection{Preliminary}
\label{sec:4.0}
Before introducing SF-NeRF, we first briefly review NeRF and its relevant methods we use for reconstructing static scene.
We adopt a conditional NeRF structure of which the emitted radiance is conditioned on an image-wise latent embedding.
Moreover, we use Integrated Positional Encoding (IPE) that featurizes a region of space instead of a single point, enabling a single MLP to learn multiscale scene representation.

\paragraph{Neural Radiance Fields (NeRF).}
NeRF~\cite{mildenhall2020nerf} represents a scene with a continuous volumetric radiance field $F_\theta$ by using a multi-layer perceptron (MLP). 
Given a 3D position $\mathbf{x}=(x,y,z) \in \mathbb{R}^3$ and a viewing direction $\mathbf{d}  \in \mathbb{S}^2$, the NeRF network $F_\theta$ returns a volume density $\sigma$ and an emitted color $\mathbf{c}=(r,g,b)$.
\begin{equation}
\label{eq:nerf_notation}
    \begin{split}
        [\sigma,\ \mathbf{z}] & = \text{MLP}_{\theta_1} \left( \gamma_\mathbf{x}(\mathbf{x}) \right), \\
        \mathbf{c} & = \text{MLP}_{\theta_2} \left( \gamma_\mathbf{d}(\mathbf{d}),\ \mathbf{z} \right),
    \end{split}
\end{equation}
where $\theta=[\theta_1,\theta_2]$ are trainable parameters of MLPs, $\gamma_\mathbf{x}(\cdot)$ and $\gamma_\mathbf{d}(\cdot)$ are the pre-defined encoding functions~\cite{mildenhall2020nerf} for spatial position and viewing direction respectively.

Consider a ray $\mathbf{r}(t)=\mathbf{o}+t\mathbf{d}$ with the ray origin $\mathbf{o}\in\mathbb{R}^3$, let us denote $\sigma(t)$ and $\mathbf{c}(t)$ as the density and color at point $\mathbf{r}(t)$ respectively.
NeRF renders the expected color $\mathbf{C}(\mathbf{r})$ by using alpha composition of densities and colors along the ray which is approximated by numerical quadrature~\cite{max1995optical} as follows:
\begin{equation}
\label{eq:nerf_render}
    \begin{split}
        \mathbf{C}(\mathbf{r}) &= \sum^K_{k=1}T(t_k)\alpha(t_k)\mathbf{c}(t_k), \\
        \text{where}~~~~\alpha(t_k) &= 1 - \exp(-\sigma(t_k)\delta_k), \\
        T(t_k) &= \prod^{k-1}_{k'=1}\left( 1-\alpha(t_{k'})\right)  \\
        &= \exp\left( -\sum^{k-1}_{k'=1} \sigma(t_{k'})\delta_{k'} \right),
    \end{split}
\end{equation}
where $\{ t_k\}^K_{k=1}$ is a set of sampled points, selected by using stratified sampling for volume rendering, and $\delta_k = t_{k+1}-t_k$ refers to the distance between the adjacent sampled points.

To increase sampling efficiency, NeRF simultaneously trains two MLPs that share the same structure: ``Coarse" and ``fine" networks where the coarse network determines the sampling points to feed into the fine network.


\paragraph{Latent Conditional NeRF.}
To synthesize views from photographs with variable illumination, previous methods~\cite{martinbrualla2020nerfw,martinbrualla2020nerfw,tancik2022blocknerf,turki2022mega} have mainly used appearance embedding $\ell^{(a)}_i\in \mathbb{R}^{n^{(a)}}$ which grants NeRF the flexibility to adjust the emitted radiance of the scene for each image $\mathcal{I}_i$.
The radiance $\mathbf{c}$ in Eq.~\eqref{eq:nerf_notation} and rendered color $\mathbf{C}(\mathbf{r})$ in Eq.~\eqref{eq:nerf_render} are replaced with image-dependent $\mathbf{c}_i$ and $\mathbf{C}_i(\mathbf{r})$  as follows:
\begin{equation}
\label{eq:nerfw_color}
    \begin{split}
    \mathbf{c}_i = \text{MLP}_{\theta_2} \left( \gamma_\mathbf{d}(\mathbf{d}),\mathbf{z}, \ell^{(a)}_i \right), \\
    \mathbf{C}_i(\mathbf{r}) = \sum^K_{k=1}T(t_k)\alpha(t_k)\mathbf{c}_i(t_k).
    \end{split}
\end{equation}
Following the framework of NeRF-W~\cite{martinbrualla2020nerfw, bojanowski2017optimizing}, we adopt the appearance embeddings, optimizing them for each input image.

\paragraph{Mip-NeRF.}
Instead of casting a single ray for each pixel like NeRF, mip-NeRF~\cite{barron2021mipnerf} casts a cone of which the radius changes as the resolution of the image changes.
Mip-NeRF changes the positional encoding scheme from encoding an infinitesimally small point to integrating within the conical frustrum (Integrated Positional Encoding) for each section of the ray.
This enables mip-NeRF to learn multiscale representations and thereby combine NeRF's coarse and fine MLPs into a single MLP.
We follow the mip-NeRF structure in this work to take advantage of the halved model capacity and scale robustness since in-the-wild photographs are taken at varying camera distances and resolutions.


\subsection{Semantic-aware Scene Decomposition}
\label{sec:4.1}
\begin{figure*}[ht]
    \centering 

    \includegraphics[width=\linewidth]{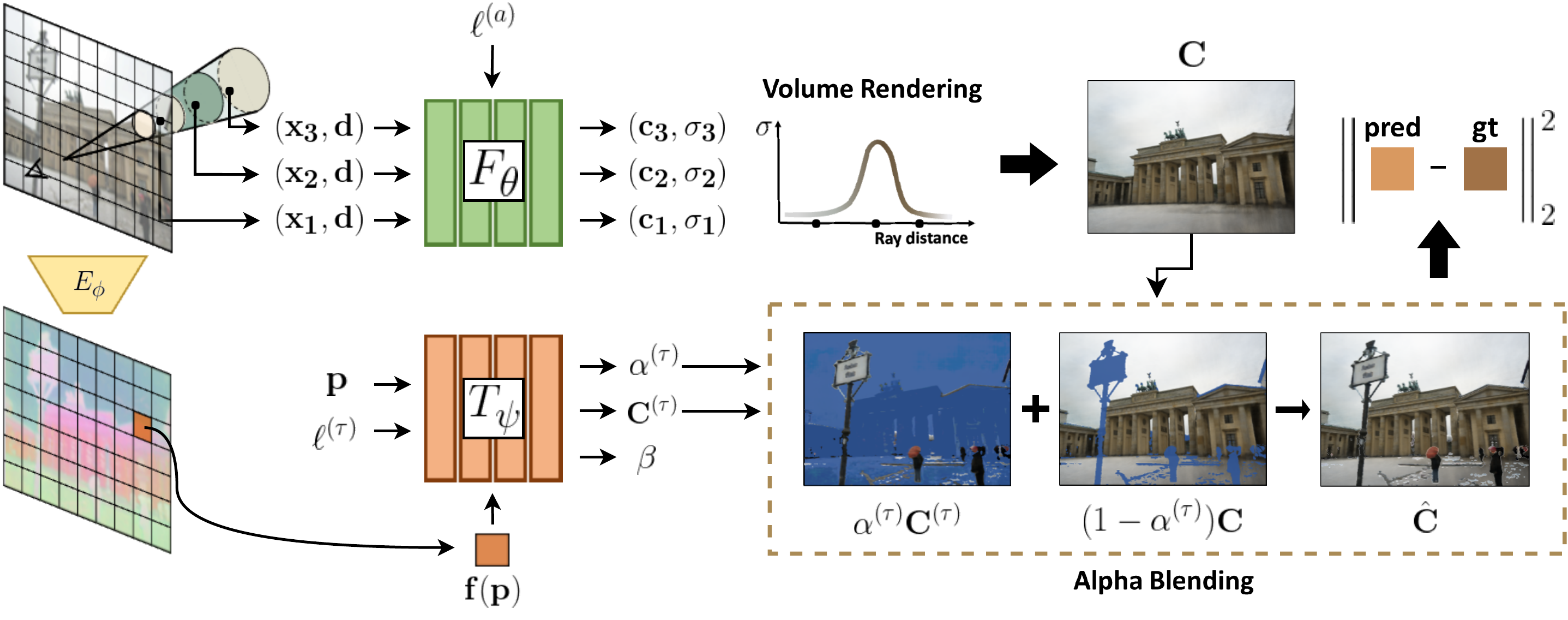} \\

    \caption{
    \textbf{Overall architecture.} Our framework decomposes the scene into static and transient components by filtering out transient occluders in the image. Given a 3D position $\mathbf{x}$, direction $\mathbf{d}$ and learned appearance embedding $\ell^{(a)}$, the static NeRF model $F_\theta$ produces static color and density, which are used to render static images. For each image, FilterNet $T_\psi$ maps pixel location $\mathbf{p}$ to its transient opacity $\alpha^{(\tau)}$, transient color $\mathbf{C}^{(\tau)}$ and uncertainty value $\beta$ conditioned on an image-dependent transient embedding $\ell^{(\tau)}$ and pixel-wise feature $\mathbf{f}(\mathbf{p})$ extracted from the image encoder $E_\phi$. The final predicted pixel color $\hat{\mathbf{C}}$ is then calculated by alpha blending rendered static color $\mathbf{C}$ and transient color $\mathbf{C}^{(\tau)}$.
    }
    \label{fig:pipeline}
\end{figure*}
To achieve consistent scene decomposition, we use an additional MLP module, dubbed FilterNet, that models transient components by leveraging semantic features of the images.  
FilterNet $T_\psi$ is designed to handle transient phenomena by learning image-dependent 2D maps: transient RGBA and uncertainty maps.
The transient RGBA image is alpha blended with the rendered (static) image produced by NeRF to reconstruct the original image as shown in Figure~\ref{fig:pipeline}.
We estimate the uncertainty of the observed color for each pixel, enabling the model to adjust its reconstruction loss by disregarding unreliable pixels.
This idea is borrowed from NeRF-W~\cite{martinbrualla2020nerfw}, but the difference is that we directly estimate a 2D uncertainty map, while NeRF-W renders the value with uncertainties of 3D locations along the ray.

To be specific, we model FilterNet $T_\psi$ as an implicit continuous function that maps a transient embedding $\ell^{(\tau)}_i \in \mathbb{R}^{n^{(\tau)}}$, a pixel location $\mathbf{p}\in \mathbb{R}^2$ and its corresponding encoded feature $\mathbf{f}_i(\mathbf{p})\in\mathbb{R}^F$ to transient color $\mathbf{C}^{(\tau)}_i$, opacity $\alpha^{(\tau)}_i\in[0,1]$ and uncertainty $\beta_i\in\mathbb{R}^+$ values as:
\begin{equation}
\label{eq:FilterNet}
    \begin{split}
        \left[\alpha^{(\tau)}_i, \mathbf{C}^{(\tau)}_i,\beta_i\right] &= T_{\psi}\hspace{-0.1cm}\left( \gamma_\mathbf{p}(\mathbf{p}), \ell^{(\tau)}_i, \mathbf{f}_i(\mathbf{p}) \right), 
    \end{split}
\end{equation}
where $\gamma_\mathbf{p}:\mathbb{R}^2 \rightarrow \mathbb{R}^{4L}$ is the positional encoding function applied to each pixel coordinate. 
Feature map $\mathbf{f}_i$ is extracted from input image $\mathcal{I}_i$ with the encoder $E:\mathbb{R}^{H\times W\times 3}\rightarrow \mathbb{R}^{H\times W \times F}$ that can be pre-trained on other real-world datasets.

Let us denote $\mathbf{C}^{(\tau)}_i\hspace{-0.05cm}(\mathbf{p}_\mathbf{r})$, $\alpha^{(\tau)}_i\hspace{-0.05cm}(\mathbf{p}_\mathbf{r})$ and $\beta_i(\mathbf{p}_\mathbf{r})$ as the transient color, opacity and uncertainty values of a pixel corresponding to ray $\mathbf{r}$, respectively.
The final predicted pixel color $\hat{\mathbf{C}}_i(\mathbf{r})$ is then obtained by combining transient color $\mathbf{C}^{(\tau)}_i\hspace{-0.05cm}(\mathbf{p}_\mathbf{r})$ and static color $\mathbf{C}_i(\mathbf{r})$ with alpha blending as follows:
\begin{equation}
\label{eq:alpha_blending}
    \begin{split}
        \hat{\mathbf{C}}_i(\mathbf{r}) = \alpha^{(\tau)}_i\hspace{-0.05cm}(\mathbf{p}_\mathbf{r})\hspace{0.05cm}\mathbf{C}^{(\tau)}_i\hspace{-0.05cm}(\mathbf{p}_\mathbf{r}) + \left( 1 - \alpha^{(\tau)}_i\hspace{-0.05cm}(\mathbf{p}_\mathbf{r}) \right)\hspace{-0.05cm}\mathbf{C}_i(\mathbf{r}).
    \end{split}
\end{equation}

For each ray $\mathbf{r}$ in image $\mathcal{I}_i$, we train FilterNet to disentangle transient components from the scene in an unsupervised manner with loss $\mathcal{L}^{(i)}_\text{t}$:
\begin{equation}
\label{eq:transient_loss}
\resizebox{.95\hsize}{!}{$
    \mathcal{L}^{(i)}_\text{t}(\mathbf{r}) \!=\! \frac{\big\lVert \hat{\mathbf{C}}_i(\mathbf{r})- \bar{\mathbf{C}}_i(\mathbf{r})\big\rVert^2_2}{2\beta_i(\mathbf{p}_\mathbf{r})^2} + \frac{\log\beta_i(\mathbf{p}_\mathbf{r})^2}{2} + \lambda_\alpha \alpha^{(\tau)}_i\hspace{-0.05cm}(\mathbf{p}_\mathbf{r}),$}
\end{equation}
where $\bar{\mathbf{C}}$ is the ground-truth color. The first and second terms can be viewed as the negative log-likelihood of $\bar{\mathbf{C}}_i(\mathbf{r})$ which is assumed to follow an isotropic normal distribution with mean $\hat{\mathbf{C}}_i(\mathbf{r})$ and variance $\beta_i(\mathbf{p}_\mathbf{r})^2$~\cite{martinbrualla2020nerfw}.
The third term discourages FilterNet from describing static phenomena.



\subsection{Transient Opacity Reparameterization}
\label{sec:4.2}
We empirically found that a naive adoption of FilterNet yields ambiguous transient opacity values, \emph{i.e}., values that are not close to zero or one, if trained to predict them by simply passing the output of the MLP through a sigmoid activation function.
This ambiguous decomposition leads to blurry artifacts in the static scene.

We encourage transient images to be either fully opaque or empty by modeling $\alpha^{(\tau)}_i$ as a Binary Concrete random variable~\cite{maddison2016concrete}, which is a continuous relaxation of Bernoulli random variable, and predicting its probability.
Binary Concrete distribution is a special case of Concrete distribution, also known as Gumbel-Softmax distribution~\cite{jang2016categorical}, which is usually used to approximate discrete random variables.

We sample $\alpha^{(\tau)}_i$ from a Binary Concrete distribution with location parameter $\tilde{\alpha}_i\in(0,\infty)$, which our FilterNet will predict instead of directly predicting the opacity value as follows:
\begin{gather}
\label{eq:binconcrete}
    \!\!\!\! \alpha^{(\tau)}_i \!= \text{sigmoid}\!\left( \frac{1}{t} \cdot  \left( \log\tilde{\alpha}_i + \log U - \log \left(1-U \right) \right) \right), \\
    U\sim\text{Uniform}(0,1),
\end{gather}
where $t\in(0,\infty)$ is a hyperparameter. 
This sampling scheme encourages our model to predict opacity values concentrated on the boundaries of the interval $[0,1]$, while allowing backpropagation.
During evaluation, we fix $U$ to $0.5$.

\subsection{Transient Opacity Smoothness Prior}
\label{sec:4.3}
The inputs of Filternet: encoded pixel (PE features) $\gamma_\mathbf{p}(\mathbf{p})$, image feature $\mathbf{f}(\mathbf{p})$, and embedding $\ell^{(\tau)}$, contain high-frequency information.
This naturally leads to high-frequency outputs which is suitable for predicting the color of a pixel, but could cause noisy prediction of transient opacity.
We thus add a smoothness loss on the transient opacity field as:
\begin{equation}
\label{eq:smooth_loss}
    \mathcal{L}^{(i)}_\text{sm}(\mathbf{p}) =  \sum^{L-1}_{k=0}2^k\bigg\lVert \frac{\partial\alpha^{(\tau)}_i\hspace{-0.05cm}(\mathbf{p})}{\partial\gamma_k(\mathbf{p})} \bigg\rVert_1,
\end{equation}
where $\gamma_k(\mathbf{p})=\left[ \cos(2^k\pi\mathbf{p}),\sin(2^k\pi\mathbf{p}) \right]$ is the $k$-th frequency encoding of $\mathbf{p}$.
The loss term multiplied by a constant value is the upper bound of L1-norm of the derivative of $\alpha^{(\tau)}_i$ with respect to the pixel coordinate, \emph{i.e}., $\Big\lVert \frac{\partial\alpha^{(\tau)}_i\hspace{-0.05cm}(\mathbf{p})}{\partial\mathbf{p}} \Big\rVert_1 \leq \sum^{L-1}_{k=0}2^k\pi\Big\lVert \frac{\partial\alpha^{(\tau)}_i\hspace{-0.05cm}(\mathbf{p})}{\partial\gamma_k(\mathbf{p})} \Big\rVert_1$, which essentially enforces smoothness on transient opacity field.
We have taken this indirect approach to reduce the sensitivity of our model to the regularization term.
The inequality above can be simply derived by using chain rule and triangle inequality.


\subsection{Optimization}
\label{sec:4.4}
We jointly optimize the model parameters $(\theta, \phi, \psi)$, the per-image appearance embeddings $\{\ell^{(a)}_i\}^N_{i=1}$ and the transient embeddings $\{\ell^{(\tau)}_i\}^N_{i=1}$ by minimizing the total loss:
\begin{equation}
\label{eq:total_loss}
\resizebox{1.\hsize}{!}{$
    \sum\limits^N_{i=1}\sum\limits_{\mathbf{r}\in\mathcal{R}_i} \!\! \big[ \mathcal{L}^{(i)}_\text{t}(\mathbf{r})+ \lambda_\text{c}\mathcal{L}^{(i)}_\text{c}(\mathbf{r}) + \lambda_\text{sm} \mathcal{L}^{(i)}_\text{sm}(\mathbf{p}_\mathbf{r})+ \lambda_\text{sp} \mathcal{L}_\text{sp}(\mathbf{r}) \big] \!+\!\lambda_a \big\lVert \ell^{(a)}_i \big\rVert^2_2,$}
\end{equation}
where $\lambda_\text{c},\lambda_\text{sm}, \lambda_\text{sp}$ and $\lambda_a$ are the hyperparameters and $\mathcal{L}^{(i)}_\text{c}$ is the reconstruction loss of rendered images composited with coarse samples $\mathbf{t}^c=\{ t^c_k \}^K_{k=1}$, and $\mathcal{L}_\text{sp}$ is a regularization term named sparsity loss:
\begin{gather}
\label{eq:coarse_sparsity_loss}
    \mathcal{L}^{(i)}_\text{c}(\mathbf{r})=\left( 1-\alpha^{(\tau)}_i(\mathbf{p}_\mathbf{r}) \right)\big\lVert \mathbf{C}_i(\mathbf{r};\mathbf{t}^c) -\bar{\mathbf{C}}_i(\mathbf{r}) \big\rVert^2_2,\\
    \mathcal{L}_\text{sp}(\mathbf{r}) = \sum_{k}\log\left( 1+ 2\sigma(t^c_k)^2 \right),
\end{gather}
where $\mathbf{t}^c=\{ t^c_k \}^K_{k=1}$ are the coarse samples produced with stratified sampling.

Sparsity loss $\mathcal{L}_\text{sp}$ encourages the static scene to have zero density on the unobserved areas.
Since transient objects hide the static scene, the static model cannot learn the scene representation on those areas.
When sufficient training images are given, this obstruction is negligible since other images provide the chance to observe the missing area.
However, in the few-shot setting, this cannot be guaranteed and allows the model to freely generate arbitrary geometry in the unobserved regions.
We hence use this sparsity prior which is also known as Cauchy loss~\cite{2021bakingnerf,yu2021plenoxels} to encourage the sparsity of NeRF's opacity field.



\section{Experiments}
\label{sec:5}
In this section, we show that our method, referred to as SF-NeRF, outperforms the previous state-of-the-art approach both quantitatively and qualitatively.
Then we provide extensive ablation studies and visualizations to validate the effectiveness of the components in SF-NeRF.

\paragraph{Datasets.}
We demonstrate our approach on four training sets which consist of internet photo collections of cultural landmarks from the Phototourism dataset~\cite{jin2021phototourism}: ``Brandenburg Gate”, ``Sacre Coeur”, ``Trevi Fountain” and ``Taj Mahal".
We follow the split used by HA-NeRF~\cite{chen2022hanerf} and downsample the images by 2 times as HA-NeRF did.
For each landmark, we evaluate our method under the few-shot settings, sampling 15 and 30 images out of 700$\sim$1700 images.

\paragraph{Baselines.}
We compare our method to NeRF~\cite{mildenhall2020nerf}, NeRF-W~\cite{martinbrualla2020nerfw}, HA-NeRF~\cite{chen2022hanerf} and one variant of NeRF-W which we call as NeRF-AM.
NeRF-AM shares the same structure as NeRF-W with its transient head eliminated (a.k.a., NeRF-A) and eliminates transient objects using a pretrained semantic segmentation model.
NeRF-AM uses DeepLabv3+~\cite{chen2018encoder} trained on Cityscapes dataset~\cite{cordts2016cityscapes} and masks out movable objects such as people (person, rider) and vehicles (rider, car, truck, bus, train, motorcycle, bicycle).
For a fair comparison, all models share the same static NeRF model architecture which consists of 8 layers of 256 hidden units for generating density $\sigma$ and one additional layer of 128 hidden units for color $\mathbf{c}$.

\paragraph{Implementation details.}
For the image encoder $E_\phi$, we use DINO~\cite{caron2021emerging} model, a pretrained self-supervised 2D image feature extractor, followed by 3 fully connected layers of 128 hidden units.
Since the feature map extracted by DINO is reduced in size, we first resize them to the original image size and then pass through the MLP layers. 
In the few-shot setting, the encoder is first pre-trained on full images of "Temple Nara Japan" in Phototourism to learn the prior of transient objects.
FilterNet $T_\psi$ consists of 5 fully connected layers of 128 hidden units, with a sigmoid and two softplus activation functions for $\mathbf{c}^{(\tau)}_i$, $\beta_i$ and $\tilde{\alpha}_i$ respectively.
For each $\beta_i$, we add $\beta_\text{min}\hspace{-0.07cm}=\hspace{-0.07cm}0.1$ to assign minimum importance to each ray.
Each appearance and transient embedding has an embedding dimension of size $n^{(a)}=48$ and $n^{(\tau)}=128$.

\paragraph{Evaluation}
All methods are evaluated on the task of novel view synthesis. 
We report the standard image quality metrics, Peak Signal to Noise Ratio (PSNR), Structural Similarity Index Measure (SSIM)~\cite{wang2004image}, and Learned Perceptual Image Patch Similarity (LPIPS)~\cite{zhang2018unreasonable} with AlexNet for all evaluations.
Since the appearance embeddings are only optimized for training set images, a procedure to obtain appearance embeddings for each test image is essential.
We evaluate each baseline following their evaluation scheme, where NeRF-W optimizes the embedding on the left half of each test image and reports metrics on the right half, and HA-NeRF obtains the appearance embedding by its learned encoder.
Here we evaluate our approach using the scheme of NeRF-W.

\begin{figure*}[ht]
    \centering 

    \includegraphics[width=\linewidth]{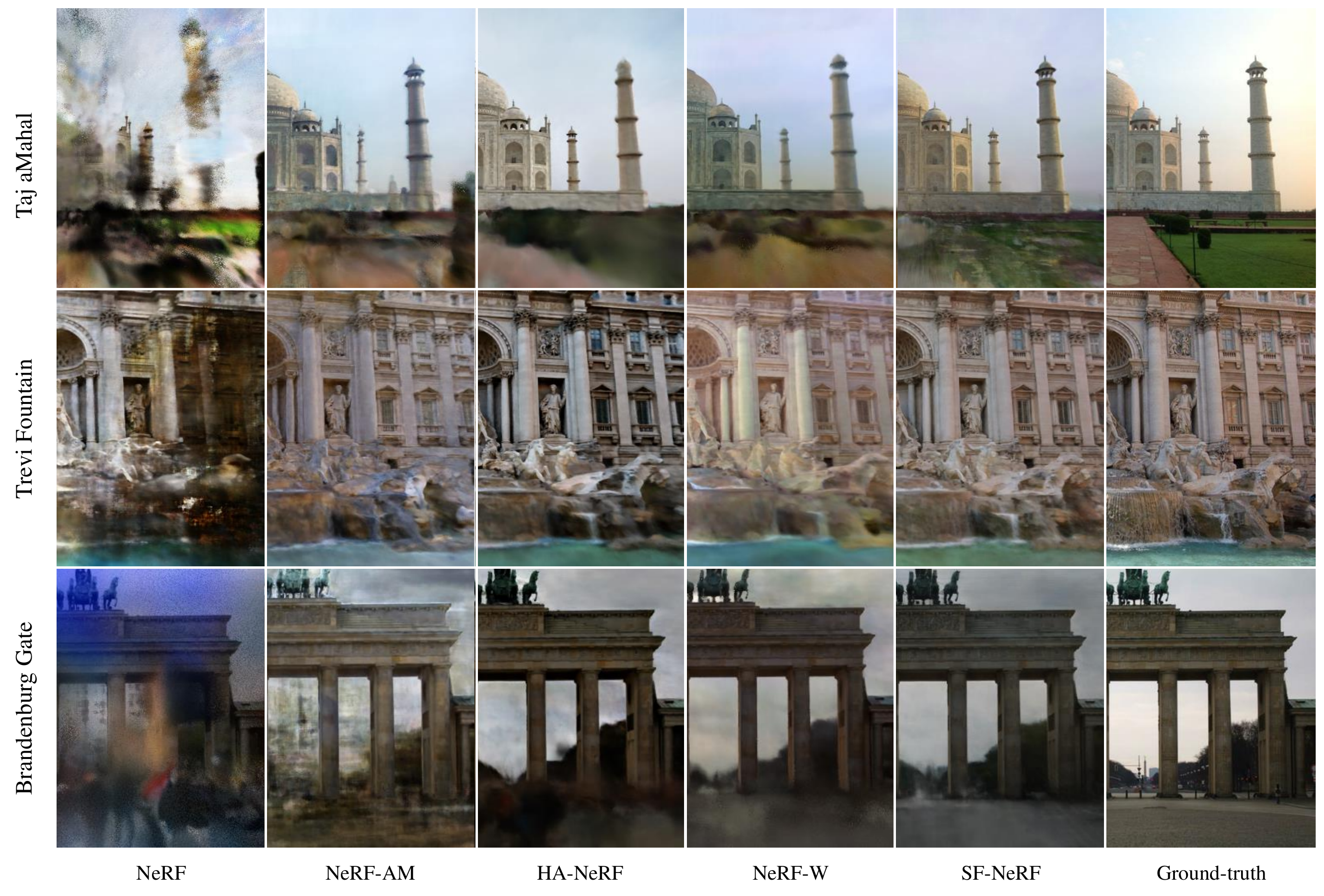} \\

    \caption{
    \textbf{Qualitative Results} of novel view synthesis on Phototourism dataset~\cite{jin2021phototourism} under the fiew-shot setting (30 images). SF-NeRF learns the separated static scene well while other baselines remain ghosting/blurry artifacts.
    }
    \label{fig:qual_fig}
\end{figure*}
\begin{figure*}[ht]
    \centering 

    \includegraphics[width=\linewidth]{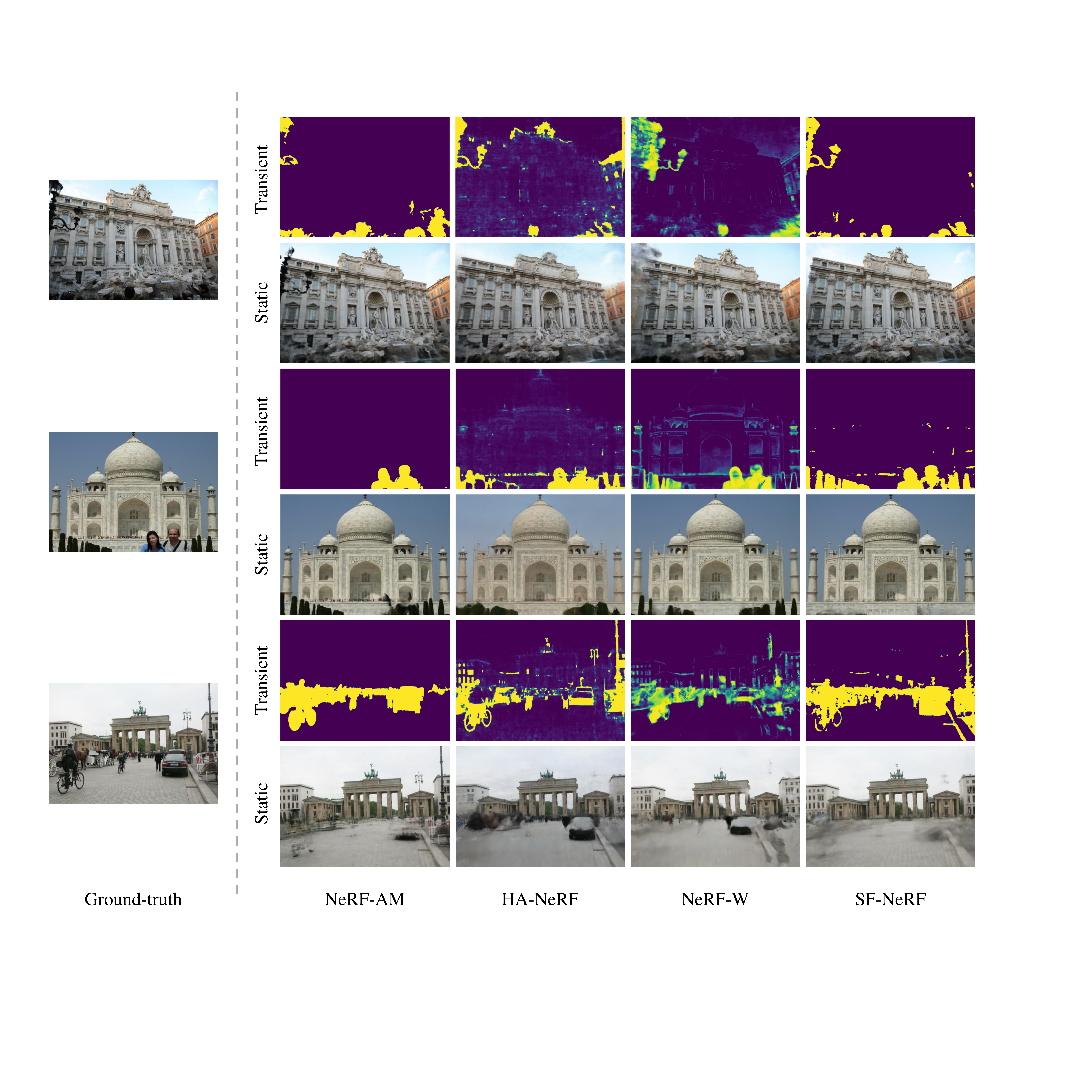} \\

    \caption{
    \textbf{Predicted transient opacity/visibility maps (Transient) and rendered static images (Static)} of SF-NeRF and other baselines. The predicted transient opacity map of NeRF-AM is based on the result of pretrained segmentation model, HA-NeRF directly learns an image-wise 2D visibility map and that of NeRF-W is rendered with the 3D transient field of NeRF-W.
    }
    \label{fig:train_qual}
\end{figure*}
\begin{table*}[ht]
\centering
\begin{adjustbox}{max width=\textwidth}
\setlength{\aboverulesep}{0pt}
\setlength{\belowrulesep}{0pt}
\begin{tabular}{lcccccccccccc}
    \hline
     & \multicolumn{3}{c}{Brandenburg Gate} & \multicolumn{3}{c}{Sacre Coeur} & \multicolumn{3}{c}{Trevi Fountain} & \multicolumn{3}{c}{Taj Mahal} \\
    \cmidrule(lr){2-4} \cmidrule(lr){5-7} \cmidrule(lr){8-10} \cmidrule(lr){11-13}
    \multirow{-2}{*}{Few15} & {PSNR}  & {SSIM}  & {LPIPS} &
                       {PSNR}  & {SSIM}  & {LPIPS} &
                       {PSNR}  & {SSIM}  & {LPIPS} &
                       {PSNR}  & {SSIM}  & {LPIPS} \\ 
    \hline
    
    NeRF~\cite{mildenhall2020nerf} 
    & 11.54 & 0.536 & 0.503 
    & 13.25 & 0.581 & 0.409
    & 12.36 & 0.415 & 0.474
    & 11.99 & 0.564 & 0.506 \\
    
    NeRF-AM 
    & 13.65 & 0.627 & 0.468 
    & 17.53 & 0.675 & 0.290 
    & 17.53 & 0.538 & 0.351 
    & 19.21 & 0.733 & 0.313  \\

    NeRF-W~\cite{martinbrualla2020nerfw}  
    & \textbf{20.80} & 0.787 & 0.266 
    & \textbf{17.87} & \textbf{0.715} & \textbf{0.235} 
    & 18.15 & 0.596 & 0.333 
    & 20.32 & 0.787 & 0.259 \\
    
    HA-NeRF~\cite{chen2022hanerf}  
    & 13.83 & 0.693 & 0.516 
    & 15.41 & 0.675 & 0.423 
    & 14.60 & 0.540 & 0.416 
    & 13.84 & 0.703 & 0.458  \\
    
    \hline
    
    SF-NeRF 
    & 19.72 & \textbf{0.790} & \textbf{0.260} 
    & 17.69 & 0.707 & 0.240 
    & \textbf{18.94} & \textbf{0.607} & \textbf{0.281} 
    & \textbf{20.47} & \textbf{0.801} & \textbf{0.226}  \\
    
    \hline
\end{tabular}
\end{adjustbox}

\vspace{+0.1cm}

\begin{adjustbox}{max width=\textwidth}
\setlength{\aboverulesep}{0pt}
\setlength{\belowrulesep}{0pt}
\begin{tabular}{lcccccccccccc}
    \hline
     & \multicolumn{3}{c}{Brandenburg Gate} & \multicolumn{3}{c}{Sacre Coeur} & \multicolumn{3}{c}{Trevi Fountain} & \multicolumn{3}{c}{Taj Mahal} \\
    \cmidrule(lr){2-4} \cmidrule(lr){5-7} \cmidrule(lr){8-10} \cmidrule(lr){11-13}
    \multirow{-2}{*}{Few30} & {PSNR}  & {SSIM}  & {LPIPS} &
                       {PSNR}  & {SSIM}  & {LPIPS} &
                       {PSNR}  & {SSIM}  & {LPIPS} &
                       {PSNR}  & {SSIM}  & {LPIPS} \\ 
    \hline
    
    NeRF~\cite{mildenhall2020nerf} 
    & 11.42 & 0.514 & 0.488 
    & 10.04 & 0.421 & 0.551
    & 11.31 & 0.333 & 0.496
    & 11.28 & 0.490 & 0.588 \\
    
    NeRF-AM 
    & 13.20 & 0.628 & 0.434
    & 12.94 & 0.506 & 0.531
    & 15.44 & 0.441 & 0.445
    & 18.41 & 0.686 & 0.338 \\

    NeRF-W~\cite{martinbrualla2020nerfw}  
    & 22.74 & \textbf{0.847} & 0.188
    & 19.31 & 0.750 & 0.205 
    & 19.54 & 0.646 & 0.298 
    & \textbf{21.03} & 0.792 & 0.259 \\
    
    HA-NeRF~\cite{chen2022hanerf}  
    & 19.88 & 0.803 & 0.278 
    & 17.66 & 0.736 & 0.256 
    & 16.97 & 0.598 & 0.352 
    & 17.92 & 0.785 & 0.321 \\
    
    \hline
    
    SF-NeRF 
    & \textbf{23.23} & 0.846 & \textbf{0.178}
    & \textbf{19.64} & \textbf{0.757} & \textbf{0.186} 
    & \textbf{20.24} & \textbf{0.657} & \textbf{0.243} 
    & 20.86 & \textbf{0.820} & \textbf{0.208} \\
    
    
    \hline
\end{tabular}
\end{adjustbox}

\vspace{0.3cm}
\caption{Quantitative results of experiments on Phototourism dataset~\cite{jin2021phototourism} in the few-shot settings (15/30 images). We report PSNR/SSIM (higher is better) and LPIPS (lower is better). SF-NeRF mostly outperforms the baselines in the few-shot setting.}
\label{tab:main_tab}
\end{table*}
\begin{figure*}[ht]
    \centering 

    \includegraphics[width=\linewidth]{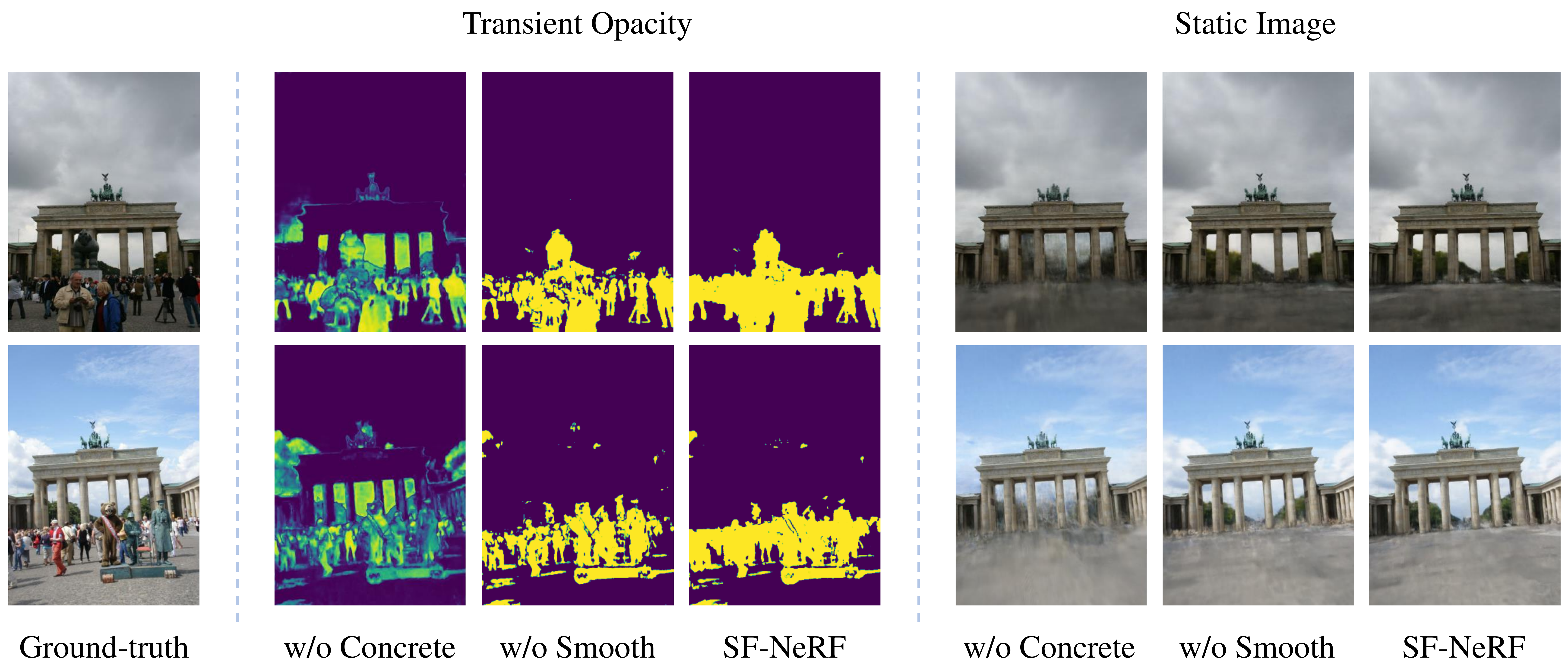} \\

    \caption{
    \textbf{Qualitative ablation results} on ``Brandenburg Gate" dataset where we compare our method with two other ablations of SF-NeRF: without transient opacity reparameterization (Concrete) or transient opacity smoothness prior (Smooth).
    }
    \label{fig:ablation_fig}
\end{figure*}
\begin{figure}[ht]
    \centering 

    \includegraphics[width=\linewidth]{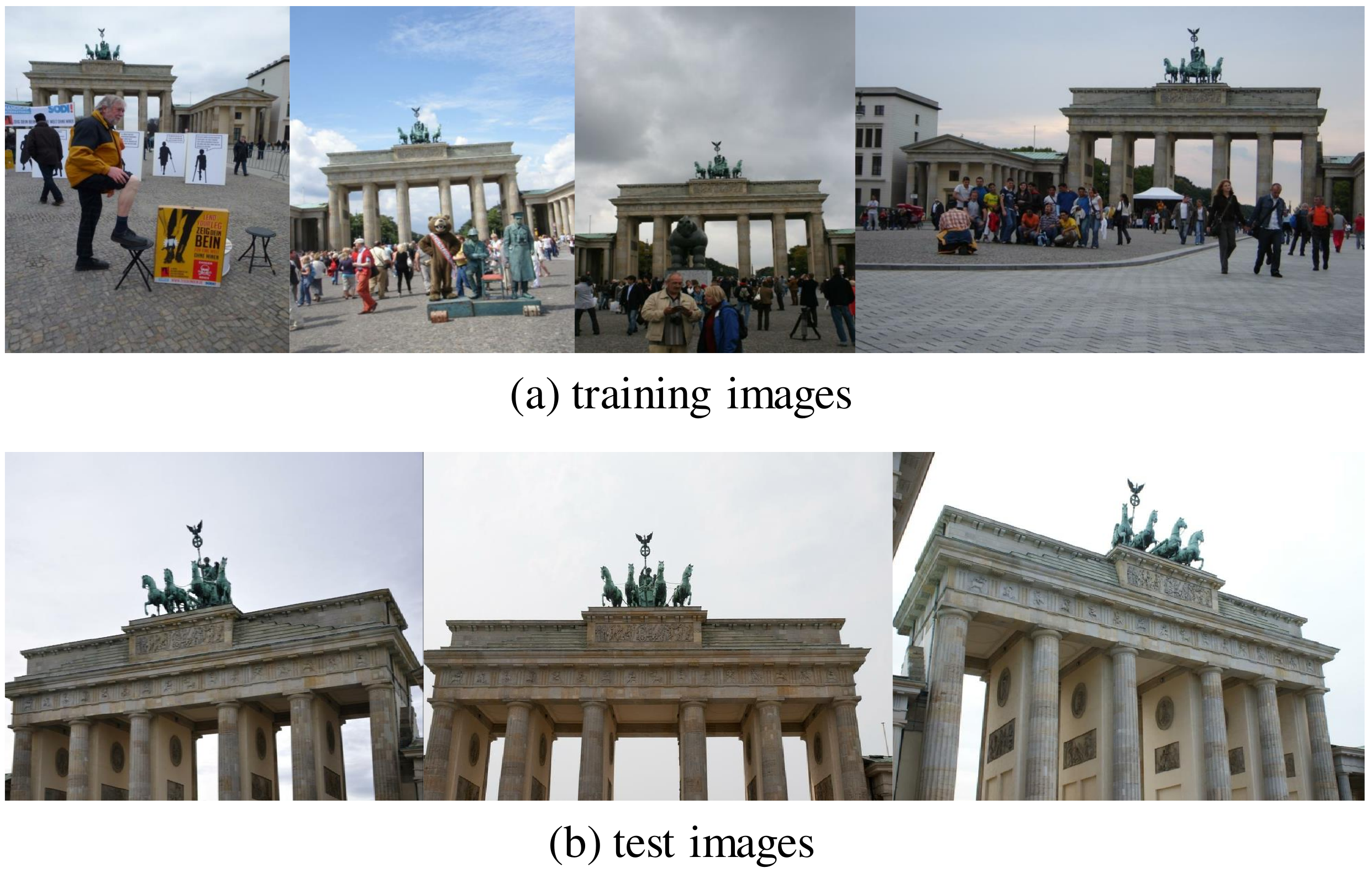} \\

    \caption{
    \textbf{Examples of training and test images from Phototourism dataset}~\cite{jin2021phototourism}. The high part of the landmark, where the transient occluders can be hardly placed, occupies a large part of the test images.
    }
    \label{fig:datas}
\end{figure}
\begin{table}[ht]
\centering
\begin{tabular}{@{\,\,}l|ccc@{\,}}
& PSNR$\uparrow$ & SSIM$\uparrow$ & LPIPS$\downarrow$  \\
\hline

w/o Concrete &  20.65 & 0.823 & 0.248  \\
w/o Smooth &  22.27 & 0.802 & 0.191  \\
SF-NeRF &  \textbf{23.23} & \textbf{0.846} & \textbf{0.178}  \\

\end{tabular}\vspace{2mm}
\caption{Ablation study of transient opacity reparameterization (Concrete) and smoothness prior (Smooth) on ``Brandenburg Gate" dataset in the few-shot setting (30 images).
}
\label{table:ablations}
\end{table}
\subsection{Results}
The overall quantitative results are shown in Table~\ref{tab:main_tab}, where SF-NeRF mostly outperforms the baselines in the few-shot setting. 
While SF-NeRF overall improves performance, the gap is not dramatic.
This is explainable if we look at the test set of Phototourism dataset.
As shown in Figure~\ref{fig:datas}, most of the test images contain only the visible part, \emph{i.e.}, the parts that were not hidden by the transient occluders, of the landmark during training.
Therefore, evaluating these images may not reflect the ability to decompose static and transient components.
Though the qualitative results show a clear improvement in SF-NeRF over the baselines.

Figure~\ref{fig:qual_fig} shows the qualitative results of our model and the baselines on some examples of the dataset.
Rendering with NeRF often results in global color shifts and ghosting artifacts.
These are the direct consequences of the assumption of NeRF we've explained in Section~\ref{sec:1} that the scene is constant across all images and every content in the scene is static.
While NeRF-AM, HA-NeRF, and NeRF-W are able to model varying photometric effects thanks to the usage of appearance embeddings, they also suffer from artifacts.
Specifically, the three baselines show ghosting artifacts in ``Taj Mahal" and ``Brandenburg Gate" (especially on NeRF-AM) and blurry artifacts in ``Trevi Fountain".
It can be seen that most of the artifacts are placed on the area that are often hidden from the transient occluders, which indicates that they lack the ability to decompose static and transient components.
This observation is supported by Figure~\ref{fig:train_qual} where NeRF-AM, HA-NeRF and NeRF-W often miss to remove some of the transient objects. 
On the contrary, SF-NeRF consistently disentangles transient elements from the static scene which proves the effectiveness of the semantic-guided filtering module.

\subsection{Ablations Studies}
We conduct ablation studies to analyze the individual contribution of each component in the proposed method.
``w/o Concrete" removes the transient opacity reparameterization trick and replaces the activation of $\tilde\alpha_i$ from softplus to sigmoid, and ``w/o Smooth" removes the transient opacity smoothness regularization term $\mathcal{L}^{(i)}_\text{sm}$.
We evaluate on the ``Brandenburg Gate" dataset and provide the results in Table~\ref{table:ablations} and Figure~\ref{fig:ablation_fig}.

As shown in Table~\ref{table:ablations}, using all the components (Concrete reparameterization and smoothness prior) achieves considerable improvement compared to the ablations of SF-NeRF: improves by on average 2.6dB and 1.0dB in PSNR compared to ``w/o Concrete" and ``w/o Smooth" respectively.
Figure~\ref{fig:ablation_fig} shows qualitative ablation results on the two components.
As mentioned in Section~\ref{sec:4.2}, if the Concrete reparameterization trick is removed (``w/o Concrete"), FilterNet yields ambiguous transient opacity values which leads to blurry, ghosting artifacts.
In the case of the transient opacity smoothness prior, ``w/o Smooth" predicts noisy transient opacity, in other words, the opacities are sparsely predicted.
The missing part of the opacities may also cause such artifacts, which is shown in Figure~\ref{fig:ablation_fig}.

\subsection{Limitations and Future Work}
SF-NeRF solely focus on removing the transient phenomena well in the way of solving the few-shot novel view synthesis from the in-the-wild photographs.
In order to improve further, approach to learn the geometry and varying appearance of the static scene well in the few-shot setting is necessary.
Furthermore, the camera parameters for each image which are obtained by using structure-from-motion~\cite{schonberger2016structure} are not entirely accurate.
Since we train with few images, SF-NeRF is sensitive to camera calibration errors and this leads to blurry reconstructions.
Therefore, simultaneously performing camera pose refinement could be one solution.
\section{Conclusion}
\label{sec:6}
We propose a novel learning framework to learn neural representations from a small number of in-the-wild photographs, named as ``\textbf{S}emantic-aware Occlusion \textbf{F}iltering \textbf{Ne}ural \textbf{R}adiance \textbf{F}ields (SF-NeRF)".
SF-NeRF focus on decomposing the transient and static phenomena, predicting the transient components of each image with an additional MLP module dubbed FilterNet.
FilterNet exploits semantic information given by an image encoder pretrained in an unsupervised manner, which is the key to achieve few-shot learning.
Moreover, we apply a reparameterization technique to prevent ambiguous decomposition and employ a smoothness prior on the transient opacity.
We shot that SF-NeRF overall outperforms state-of-the-art novel view synthesis methods on Phototourism dataset under the few-shot setting.

{\small
\bibliographystyle{unsrt}
\bibliography{egbib}
}

\end{document}